\documentclass[11pt,a4paper]{article}
\usepackage[table,dvipsnames]{xcolor}

\usepackage[hyperref]{emnlp2020}
\usepackage{url}

\usepackage{times}
\usepackage{latexsym}

\usepackage{microtype}

\usepackage{graphicx}               
\usepackage{tabularx}               
\newcolumntype{C}{>{\centering\arraybackslash}X}
\usepackage{multirow}               
\usepackage{diagbox}                
\usepackage{hhline}                 
\usepackage{color}                  
\usepackage{amsmath}                
\usepackage{bm}
\usepackage{amssymb}                
\usepackage{mathtools}              
\usepackage{enumitem}               
\usepackage{booktabs}
\usepackage{subcaption}
\usepackage{amssymb}

\usepackage{xparse}
\NewDocumentCommand{\heng}{ mO{} }{\textcolor{OrangeRed}{\textsuperscript{\textit{Heng}}\textsf{\textbf{\small[#1]}}}}
\NewDocumentCommand{\qianying}{ mO{} }{\textcolor{CadetBlue}{\textsuperscript{\textit{Qianying}}\textsf{\textbf{\small[#1]}}}}
\NewDocumentCommand{\fei}{ mO{} }{\textcolor{Blue}{\textsuperscript{\textit{Fei}}\textsf{\textbf{\small[#1]}}}}
\NewDocumentCommand{\lingfei}{ mO{} }{\textcolor{Cyan}{\textsuperscript{\textit{Lingfei}}\textsf{\textbf{\small[#1]}}}}
\NewDocumentCommand{\haoran}{ mO{} }{\textcolor{Brown}{\textsuperscript{\textit{Haoran}}\textsf{\textbf{\small[#1]}}}}

\NewDocumentCommand{\aysa}{ mO{} }{\textcolor{Purple}{\textsuperscript{\textit{Aysa}}\textsf{\textbf{\small[#1]}}}}

\aclfinalcopy
\title{Minimize Exposure Bias of Seq2Seq Models in Joint Entity and Relation Extraction}

\author{Ranran Haoran Zhang\thanks{\quad This denotes equal contribution.}$\ \,^1$, Qianying Liu\footnotemark[1]$\ \,^{2}$, Aysa Xuemo Fan$^1$, Heng Ji$^1$, Daojian Zeng$^4$, \\ \textbf{Fei Cheng$^2$, Daisuke Kawahara$^3$ and Sadao Kurohashi$^2$}\\
$^1$ University of Illinois at Urbana-Champaign\\
$^2$ Graduate School of Informatics, Kyoto University \\
$^3$ School of Fundamental Science and Engineering, Waseda University\\
$^4$ Hunan Normal University\\
{\tt \{haoranz6,xuemof2,hengji\}@illinois.edu; ying@nlp.ist.i.kyoto-u.ac.jp}\\{\tt  \{feicheng,kuro\}@i.kyoto-u.ac.jp; dkw@waseda.jp; zengdj916@163.com}
}

\date{}

\begin{document}

\maketitle

\begin{abstract}


    Joint entity and relation extraction aims to extract relation triplets from plain text directly. 
    Prior work leverages Sequence-to-Sequence (Seq2Seq) models for triplet sequence generation.
    However, Seq2Seq enforces an unnecessary order on the unordered triplets and involves a large decoding length associated with error accumulation. These methods introduce exposure bias, which may cause the models overfit to the frequent label combination, thus limiting the generalization ability. 
    We propose a novel Sequence-to-Unordered-Multi-Tree (Seq2UMTree) model to minimize the effects of exposure bias by limiting the decoding length to three within a triplet and removing the order among triplets. 
    We evaluate our model on two datasets, DuIE and NYT, and systematically study how exposure bias alters the performance of Seq2Seq models. 
    Experiments show that the state-of-the-art Seq2Seq model overfits to both datasets while Seq2UMTree shows significantly better generalization. 
    Our code is available at \url{https://github.com/WindChimeRan/OpenJERE}.


\end{abstract}

\section{Introduction}




Relation extraction aims to extract entity-relation triplets ($h, r, t$) from plain text.  For example, in the triplet \textit{(Obama, graduate\_from, Columbia University)}, \textit{Obama} and \textit{Columbia University} are the head and tail entities appearing in the text, and \textit{graduate\_from} is the relation between these two entities.
For supervised relation extraction, early studies focus on pipeline methods, which use an entity extractor 
to extract entities, and then classify the relations of entity pairs. 
These methods ignore the intrinsic interactions between these two subtasks and propagate classification errors through the tasks.
Jointly entity and relation extraction (\textsc{JERE}) considers the subtask interaction
\cite{Roth2004,Ji2005a,Ji2005b,yu-lam-2010-jointly,riedel2010modeling,Sil2013,li2014constructing,li-ji-2014-incremental,Durrett2014,miwa-sasaki-2014-modeling,lu-roth-2015-joint,yang-mitchell-2016-joint,kirschnick-etal-2016-jedi,miwa-bansal-2016-end,gupta-etal-2016-table,limb}
, but they mainly exploit feature-based system or multi-task neural network, which can not capture inter-triplet dependency.

\begin{figure*}[t]
  \centering
  \includegraphics[width=\textwidth]{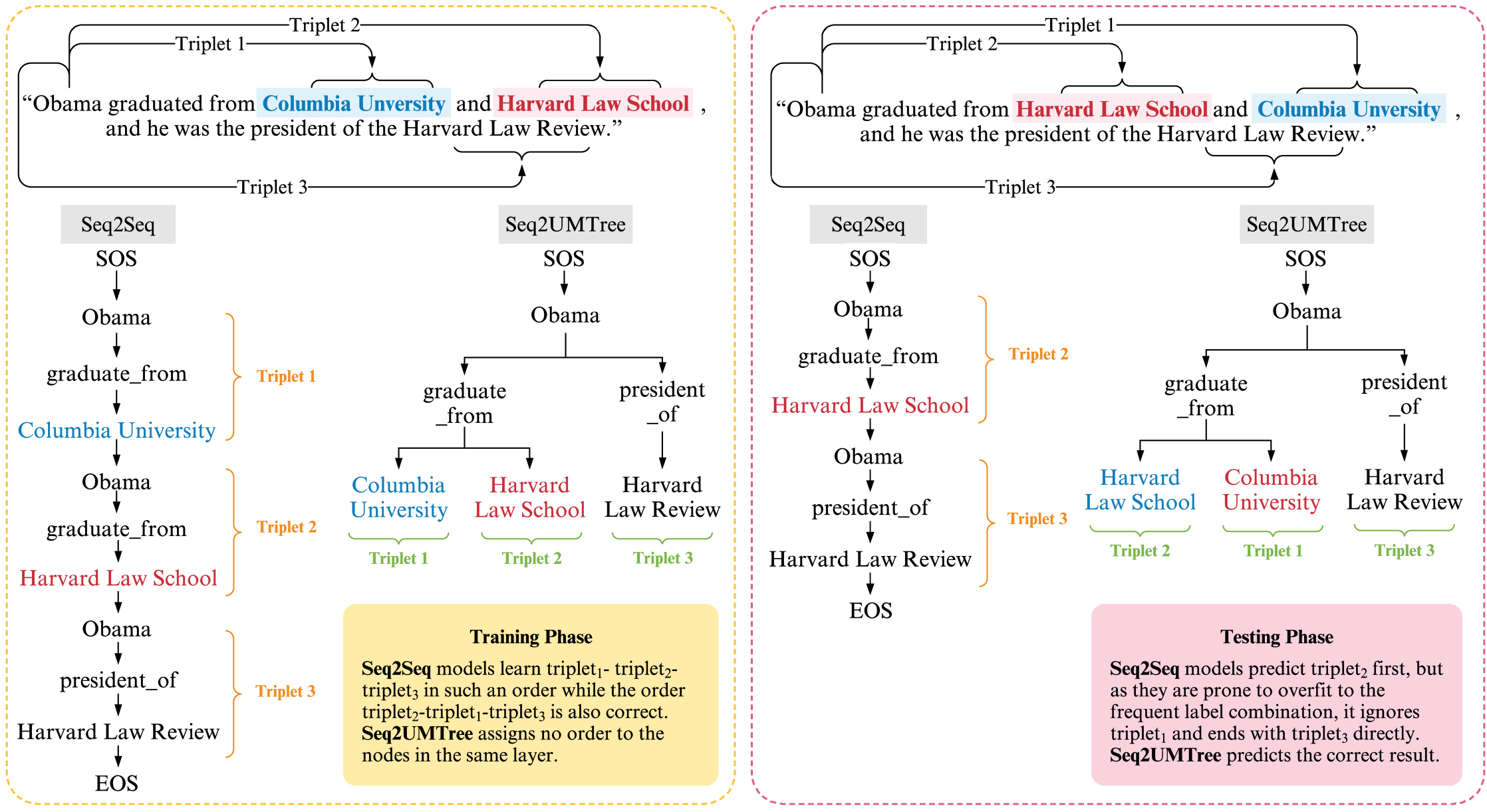}
  \caption{The training and testing of Seq2Seq and Seq2UMTree for different triplet orders.}
  \label{fig:intro}
  \end{figure*}

NovelTagging \cite{zheng-etal-2017-joint} 
integrates these two subtasks into one sequence labeling process, which assigns a single  entity-relation tag to each token; when a token belongs to multiple relations, the prediction results will be incomplete.
Instead of sequence labeling,
Sequence-to-Sequence (Seq2Seq) models \cite{cho-etal-2014-learning} are able to extract an entity multiple times, thus multiple relations can be assigned to one entity, which solves the problem naturally \cite{zeng-etal-2018-extracting,zeng2019copymtl,zeng-etal-2019-learning,nayak2019effective}. 
Specifically, all existing Seq2Seq models pre-define a sequential order 
for the target triplets, e.g. triplet alphabetical order, and then decode the triplet sequence according to the order autoregressively, which means the current triplet prediction relies on the previous output. For exmaple, in Figure \ref{fig:intro}, the triplet list is flattened to \textit{[Obama]-[graduate\_from]-[Columbia University]-[Obama]-[graduate\_from]-[Harvard Law School]...}


However, the autoregressive decoding of the Seq2Seq models introduces exposure bias problem which may severely reduce the performance. 
Exposure bias refers to the discrepancy between training and testing phases of the decoding process \cite{ranzato2015sequence}.
In the training phase, the current triplet prediction relies on the gold-standard labels of the previous triplets,
while in the testing phase, the current triplet prediction relies on the model prediction of the previous triplets, which can be different from the gold-standard labels. 
As a result, in the test phase, a skewed prediction will further deviate the predictions of the follow-up triplets; if the decoding length is large, the discrepancy from the gold-standard labels would be further accumulated. Such accumulated discrepancy may decrease the performance especially in predicting longer sequences, i.e., multi-triplet prediction. 


Furthermore, because Seq2Seq model sequentially predicts the triplets, it enforces an unnecessary order on the unordered labels, while other triplet orders are also correct. Thus, the assigned order makes the model prone to memorize and overfit to the frequent label combinations in the training set and poorly generalize to the unseen orders. The overfitting is also the side effect of exposure bias \cite{tsai2019order}, which may result in missing triplets in Seq2Seq prediction.
For example, in Figure \ref{fig:intro}, during the training phase, the Seq2Seq model learns $\text{triplet}_1$-$\text{triplet}_2$-$\text{triplet}_3$ in such an order while the order $\text{triplet}_2$-$\text{triplet}_1$-$\text{triplet}_3$ is also correct. 
In the testing phase, the Seq2Seq model predicts $\text{triplet}_2$ first based on the assigned order, but because $\text{triplet}_2$-$\text{triplet}_3$ is a frequent order for  the model, it ignores $\text{triplet}_1$ and ends with $\text{triplet}_3$ directly (i.e.,$\text{triplet}_2$-$\text{triplet}_3$).
When an order is enforced on the model, the model proceeds with more learning constrains.

To mitigate the exposure bias problem while keeping the simplicity of Seq2Seq,
we recast the one-dimension 
triplet sequence 
to two-dimension Unordered-Multi-Tree (UMTree) and propose a novel model \textbf{Seq2UMTree}.
{The Seq2UMTree model} is based on an {Encoder-Decoder} framework,
which is composed of a conventional encoder and a UMTree decoder.
The UMTree decoder
models entities and relations jointly and structurally, using a copy mechanism with unordered multi-label classification as the output layer.
This multi-label classification model ensures the nodes in the same layer are unordered and discards the predefined triplet order so that the prediction deviation will not aggregate and affect other triplets.
Different from the standard Seq2Tree \cite{dong-lapata-2016-language, liu-etal-2019-tree}, the decoding length is limited to three (one triplet), which is the shortest feasible length for \textsc{JERE} task. In this way, the exposure bias is minimized 
under the triplet-level F1 metrics. 


{In conclusion, our contributions are listed as follows:}
\begin{itemize}
    \item We point out the redundancy of the predefined triplet order of the Seq2Seq model, and propose a novel Seq2UMTree model to minimize exposure bias by recasting the ordered triplet sequence to an Unordered-Multi-Tree format.
    
    \item We systematically analyze how exposure bias diminishes the reliability of the performance scores of the standard Seq2Seq models.
\end{itemize}








  \begin{figure*}[t]
  \centering
  \includegraphics[width=\textwidth]{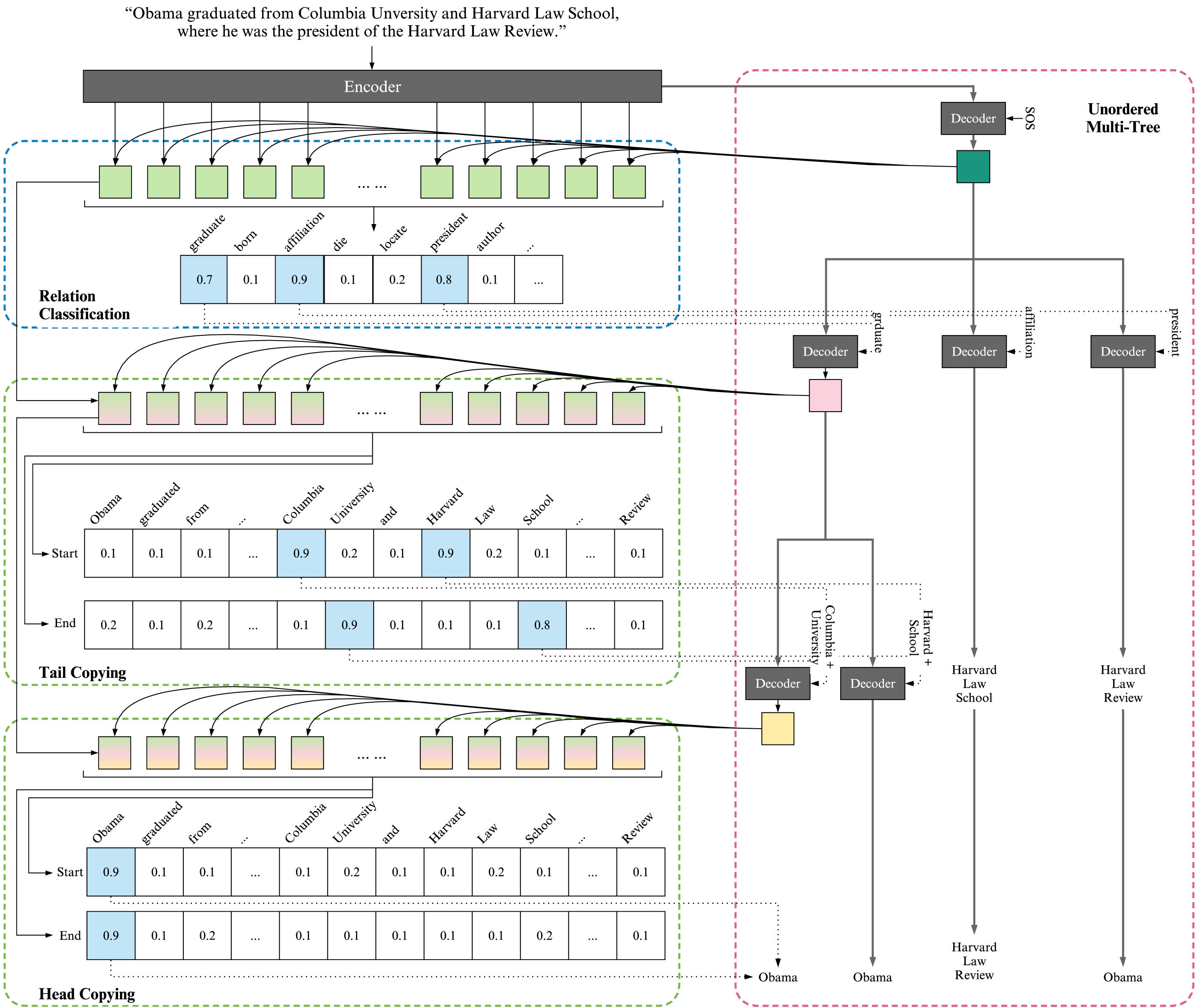}
  \caption{The model overview. The decoding order within a triplet is $r, t, h$. The relation is predicted from a predefined relation dictionary and the entities are copied from the sentence.}
  \label{fig:model}
  \end{figure*}


\section{Methodology}
The Seq2UMTree model consists of a conventional Seq2Seq encoder and a UMTree decoder. 
The UMTree decoder is different from the standard decoder in a way that it generates unordered multi-label outputs and uses the UMTree decoding strategy.
The overview of the model is shown in Figure. \ref{fig:model}. We illustrate the model details in the following subsections.

\subsection{Model}

\begin{figure*}[ht]
\begin{subfigure}{.5\textwidth}
  \centering
  \includegraphics[width=.95\linewidth]{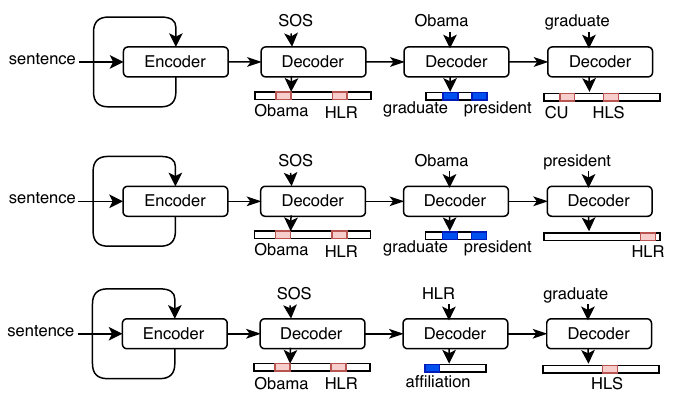}  
  \caption{Training}
  \label{fig:sub-first}
\end{subfigure}
\begin{subfigure}{.5\textwidth}
  \centering
  \includegraphics[width=.95\linewidth]{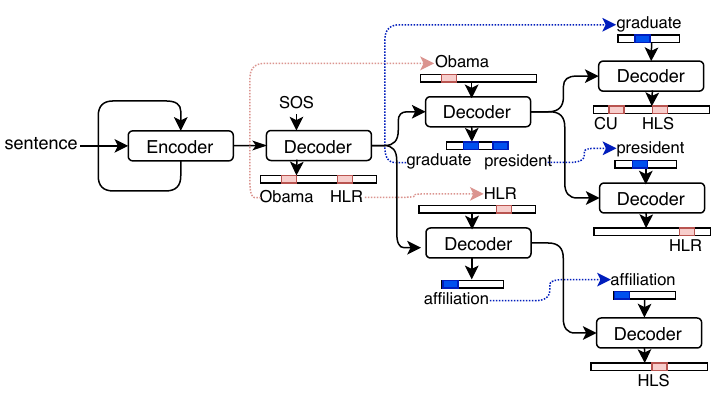}  
  \caption{Testing}
  \label{fig:sub-second}
\end{subfigure}
\caption{Seq2UMTree is trained in a teacher-forcing way by aligning the tree to the sequences. In the test phase, the model decodes the whole tree autoregressively.
In the figure, HLR, HLS, CU are the abbreviations of Harvard Law Review, Harvard Law School and Columbia University. The example uses $h$-$r$-$t$ as the order within a triplet.}
\label{fig:train_test}
\end{figure*}

Formally, the input sentence $\bm{x} = [x_0, x_1,\dots, x_n]$ is first transformed to a sequence of context aware representations by word embedding and Bidirectional
Recurrent Neural Network (Bi-RNN) \cite{birnn} with Long Short Term Memory (LSTM) \cite{lstm} as the encoder:

\begin{equation}
    [\bm{s}^E_0, \bm{s}^E_1,\dots, \bm{s}^E_n] = \text{Encoder}([x_0, x_1,\dots, x_n])
\end{equation}
Then we pass the output $\bm{s}$ sequence to $\text{Conv}_{en}$:
\begin{equation}
    \bm{o}_0 = \text{Conv}_{en}([\bm{s}^E_0, \bm{s}^E_1,\dots, \bm{s}^E_n])
\end{equation}
 where $\text{Conv}_{en}$ is the encoder convolutional layer.  $\text{Conv}_{en}$ maps $\bm{s}^E$ to $\bm{o}_0$, which is also a sequence and has the identical dimension as the $\bm{s}$ sequence. The output is denoted as $\bm{o}_0 \in \mathbb{R}^{n \times h}$, where $h$ is the hidden size, $n$ is the length of the input sentence. $\bm{o}_0$ is the auxiliary representation of the sentence, which is used for decoding with scratchpad attention mechanism \cite{benmalek-etal-2019-keeping}:  $\bm{o}_{n-1}$ is used to calculate attention score, and $\bm{o}_{n-1}$ will be updated to $\bm{o}_{n}$ at every decoding step.

During decoding, we use different input embeddings and output layers for relation and entity extraction, and they share the same decoder parameters.
For the input embedding $\bm{w}_t$, we use: (a) ``start-of-the-sentence'' embedding: $\bm{w}_0^{sos} \in \mathbb{R}^h$, which is always the beginning of the decoding and is considered as depth 0, (b) relation embedding: $\bm{w}^r_t \in \mathbb{R}^h$, (c) entity embedding:
$\bm{w}^e_t = \bm{o}_{t-1}^{e1} + \bm{o}_{t-1}^{e2} \in \mathbb{R}^h$,
where $e1$ and $e2$ are the beginning position and the end position of the predicted entity respectively. $t \in \{1,2,3\}$, which is the decoding time step. 
The decoding order can be predefined arbitrarily, such as $h$-$r$-$t$ or $t$-$r$-$h$.

Given the input embedding $\bm{w}_t$ and the output of the previous time step $\bm{s}^D_{t-1}$, a unary LSTM decoder is used to generate decoder hidden state:

\begin{equation}
    \bm{s}_t^D = \text{Decoder}(\bm{w}_t,\bm{s}_{t-1}^D)
\end{equation}
where $\bm{s}^D_{t}$ is the decoder hidden states; $\bm{s}^D_{0}$ is initialized by $\bm{s}^E_{n}$.

Attention mechanism \cite{luong-etal-2015-effective} is used to generate context-aware embedding:

\begin{equation}
    \bm{a}_{t} = \text{Attention}(\bm{o}_{t-1},\bm{s}_t^D)
\end{equation}
where $a \in \mathbb{R}^{h}$.
Then the context-aware representation $\bm{a}_{t}$ is concatenated with the original $\bm{o}_{t-1}$, followed by a convolution layer:

\begin{equation}
    \bm{o}_{t} = \text{Conv}_{de}([\bm{a}_{t};\bm{o}_{t-1}^{0:n}])
\end{equation}
where $\text{Conv}_{de}$ maps dimension $2h$ to $h$ and $\bm{a}_{t}$ is replicated $n$ times before concatenation.




The output layer of the relation prediction is a linear transformation followed by a max-pooling over sequence:

\begin{equation}
    \bm{prob}_r = \sigma(\text{Max}(\bm{o}_{t}\bm{W}_r+\bm{b}_r))
\end{equation}
where $\sigma$ is the sigmoid function for multi-relation classification, $\bm{W}_r \in \mathbb{R}^{h \times r}$, $\bm{b}_r \in \mathbb{R}^{r}$ and $\bm{prob}_r \in \mathbb{R}^{r}$ is the predicted probability vector of the relations.

The output layers of the entity prediction are two binary classification layers over the whole sequence, predicting the positions of the beginning and the end of the entities respectively:

\begin{equation}
\begin{split}
    \bm{prob}_{e_{b}} &= \sigma(\bm{W}_{e_{b}}^T\bm{o}_{t}+{b}_{e_{b}}) \\
    \bm{prob}_{e_{e}} &= \sigma(\bm{W}_{e_{e}}^T\bm{o}_{t}+{b}_{e_{e}})
    \end{split}
\end{equation}
where $\bm{W}_e \in \mathbb{R}^{h \times 1}$, ${b}_e$ is a scalar and  $\bm{prob}_e \in \mathbb{R}^{n \times 1}$ is the predicted probability vector of the entities, $e_{b}$ and $e_{e}$ refer to the beginning and the ending of the entity. Different from \newcite{nayak2019effective}, the sigmoid function $\sigma$ enables the model to predict multiple entities at a time.

\subsection{Training and Testing}

\begin{table*}[t]
\normalsize
\centering
\begin{tabular}{l|lccclccc}
\hline
\multicolumn{1}{c|}{\multirow{2}{*}{}} & \multicolumn{4}{c}{NYT}                                                                         & \multicolumn{4}{c}{DuIE}                                                   \\ \cline{2-9} 
\multicolumn{1}{c|}{}                  & \multicolumn{1}{c}{test\#} & Prec          & Rec           & \multicolumn{1}{c|}{F1}            & \multicolumn{1}{c}{test\#} & Prec          & Rec           & F1            \\ \hline
CopyMTL                                & .978                        & .685 & .648           & \multicolumn{1}{c|}{.666}           & .962                        & .496           & .394           & 439           \\
WDec                                   & .988                       & \textbf{.843} & \textbf{.764} & \multicolumn{1}{c|}{\textbf{.802}} & .919                       & .641          & .542          & .587          \\
MHS                                    & .995                       & .798          & .739          & \multicolumn{1}{c|}{.768}          & .984                       & \textbf{.772} & .623          & .690          \\
Seq2UMTree                             & 1.00                      & .791          & {.751}           & \multicolumn{1}{c|}{.771}          & 1.00                       & .756          & \textbf{.730} & \textbf{.743} \\ \hline
\end{tabular}
\caption{Main Results on NYT and DuIE. \#test is the valid sentence percentage of the test set to the models.}
\label{tab:main}
\end{table*}

In the training phase, for each sentence, 
we reorganize the training data that each pair of depth 1 and 2 (e.g. $h$-$r$) in UMTree would form one training example, so that this strategy traverses the whole tree. The training process of each node  corresponds to one time step in Seq2Seq models. We then train the model in teacher forcing \cite{6795228} manner: the input of each decoding time step is given by the gold-standard labels.
Take the order $h$-$r$-$t$ as an example, in Fig. \ref{fig:sub-first}, the total loss is the sum of the losses of the following three decoding steps:
\begin{equation}
\begin{split}
    L =& -\log \Pr (h_{b}=h_{b}^*|\bm{x};\theta) \\
    & -\log \Pr (h_{e}=h_{e}^*|\bm{x};\theta) \\
    &- \log \Pr (r=r^*|h_{b}^*, h_{e}^*, \bm{x};\theta) \\
    &- \log \Pr (t_{b}=t_{b}^*|r^*, h_{b}^*, h_{e}^*, \bm{x};\theta) \\
    &- \log \Pr (t_{e}=t_{e}^*|r^*, h_{b}^*, h_{e}^*, \bm{x};\theta)
\end{split}
\end{equation}
where $h^*, r^*, t^*$ are the ground truth of the triplets, $\theta$ is all of the trainable parameters in the model. 
In the testing phase, the UMTree uses auto-regressive decoding strategy. The decoder predicts the nodes layer by layer, where the prediction results of the previous layer are used as the input of the next time step separately, as shown in Fig. \ref{fig:sub-second}.

\section{Experiments}

\begin{table}[t]
\normalsize
\centering
\resizebox{0.48\textwidth}{!}{
\begin{tabular}{l|cc|cc}
\hline
      & \multicolumn{2}{c|}{NYT} & \multicolumn{2}{c}{DuIE} \\ \cline{2-5} 
      & \#sentence    & \#triplet   & \#sentence    & \#triplet   \\ \hline
train & 56,195      & 90,967     & 155,931     & 314,996    \\
dev   & 5,000       & 8,153      & 17,178      & 34,270     \\
test  & 5,000       & 8,214      & 21,639      & 43,749     \\ \hline
\end{tabular}
}
\caption{Data statistics of NYT and DuIE datasets. NYT contains 24 relations and DuIE contains 49 relations.}
\label{tab:whole_data}
\end{table}

\subsection{Settings}
\noindent \textbf{Dataset} \\
We evaluate our model on two datasets, NYT and  DuIE\footnote{\url{https://ai.baidu.com/broad/introduction?dataset=dureader}}.
NYT \cite{riedel2010modeling} is a English news dataset that is generated by distant supervision without manual annotation, which is widely used in \textsc{JERE} studies \cite{zheng-etal-2017-joint,zeng-etal-2018-extracting,rlre,tag2table,fu-etal-2019-graphrel,nayak2019effective,zeng2019copymtl,zeng-etal-2019-learning,chen-etal-2019-mrmep,wei2019novel}. 
We use the same data split as CopyRE \cite{zeng-etal-2018-extracting}.
DuIE \cite{li2019duie} is a large-scale Chinese \textsc{JERE} dataset where sentences are from Baidu News Feeds and Baidu Baike. 
The whole dataset is annotated by distant supervision and then checked manually. 
We take 10\% of the training set randomly as a validation set and the original development set as the test set because the original test set is not released. 
In prerprocessing, for both datasets, we filter out the sentences that contain no triplet. 
The data statistics of these two datasets are shown in Table \ref{tab:whole_data}. 


\noindent \textbf{Baselines} \\
We compare the proposed model, Seq2UMTree, with strong baselines under the same hyperparameters, as follows: 
1) CopyMTL \cite{zeng2019copymtl} is a Seq2Seq model with copy mechanism, and the entities are found by multi-task learning.
2) WDec \cite{nayak2019effective} is a standard Seq2Seq model with dynamic masking, and decode the entity token by token. 
3) MHS \cite{bekoulis2018joint} is a non-Seq2Seq baseline, which enumerates all possible token pairs.
4) Seq2UMTree is the proposed method, which generates triplets in a concise tree structure. 

\noindent \textbf{Hyperparameters} \\
For the sake of fair comparison, 
we reproduce all the baselines ourselves with the same hyperparameter settings. 
We use 200-dimension word embedding for English and character embedding for Chinese. Both are initialized from Gaussian distribution $\mathcal{N}(0,1)$, and 200-dimension Bi-LSTM encoder is used for both to mitigate the heterogeneity of these two languages.
These models are trained for 50 epochs by Adam optimizer \cite{kingma2014method}, and the models with the highest validation F1 scores are used for testing. 
The training of all compared models can be finished in 24 hours in a single NVIDIA V100 16GB GPU. 
The decoding order of Seq2UMTree in both datasets is $r$-$t$-$h$. We will discuss the effect of the order in subsection \ref{sec:order}.




     \begin{figure}[t]
  \centering
  \includegraphics[width=0.49\textwidth]{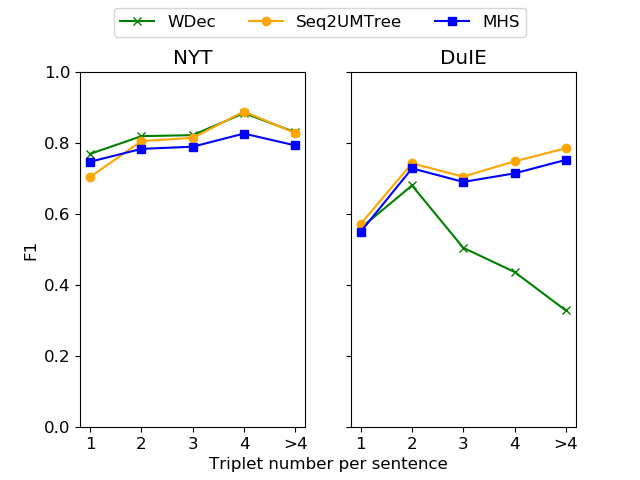}
  \caption{The F1 scores of the models on test subsets NYT and DuIE with different numbers of triplets. 
  The subsets contain sentences with number of triplets 1, 2, 3, 4 and $>$4 have 3080, 1127, 298, 315 and 470 in NYT and 9853, 7034, 2366, 1153 and 1233 in DuIE.}
  \label{fig:multi}
  \end{figure}

\subsection{Main Results}


The experiment results are shown in Table \ref{tab:main}. 
Because of the limitation of GPU memory, the Seq2Seq models and MHS cannot process all the testing data. 
The valid sentence percentage of the test set is shown in the \#test column. 
WDec sets the maximal decoding length to 50 and CopyMTL can only decode 5 triplets at most, resulting in their incomplete coverage on DuIE testset, in which 8.1\% and 3.8\% of the test sentences are deleted in the preprocessing stage. Moreover, because the entities in DuIE usually have more tokens than NYT does, the maximal decoding length of WDec filters out more examples in DuIE (8.1\%) than in NYT (1.2\%). 
MHS extracts triplets by exhaustively enumerating all token pairs,
resulting in a $\mathcal{O}(l^2r)$ GPU memory consumption of encoding sentences, where $l$ is the sentence length and $r$ is the number of relations. In our reproduction, we delete sentences longer than 100 tokens in NYT and 150 in DuIE, which covers 0.5\% of the NYT test set and 1.6\% of the DuIE test set.
Among all the models, only Seq2UMTree can be applied for all sentences in both datasets\footnote{The performance scores are calculated in their processed test sets.} and the space complexity is $\mathcal{O}(2l+r)$. 

From the Table \ref{tab:main} we can see that Seq2UMTree outperforms the previous best Seq2Seq model WDec by 15.6\% F1 score in DuIE, but it underperforms WDec in NYT by 3.1\%. The inconsistency of the performances on two datasets motivates us to conduct deeper investigation in the next section.
\section{Investigation on Data \& Model Bias}
\subsection{Exposure Bias and Generalization}
\label{sec:recitation}
While Seq2Seq assigns an order to the triplets, Seq2UMTree generates triplets in an unordered way, regardless of the triplet number.
To verify the effectiveness of Seq2UMTree on multiple triplets, 
we split NYT and DuIE test sets into five subsets in which each sentence only contains a specific number of triplets (1, 2, 3, 4, $>$4).
The performance of the models in the subsets is shown in Figure. \ref{fig:multi}.
In DuIE, when the triplet number increases, the F1 scores of WDec decrease drastically from 70\% to 40\% for triplet numbers greater than 2. MHS and Seq2UMTree perform better as the triplet number increases.
By contrast, in NYT, all models perform similarly with different numbers of triplets. To better address the reasons behind the performance differences, we conduct qualitative analysis of the data, finding that in NYT, 90\% triplets in the test set reoccurred in the training set, while in DuIE, the percentage is only 30\%.
Based on this observation, we hypothesize that the Seq2Seq models gain high score in NYT because of exposure bias: as the triplets in the test set are highly overlapped with those in the training set, the models achieve high scores by memorizing the frequently reoccurred training set triplets, which causes the overfitting that makes the models generalize poorly to the unseen triplets.

  \begin{figure}[t]
  \centering
  \includegraphics[width=0.49\textwidth]{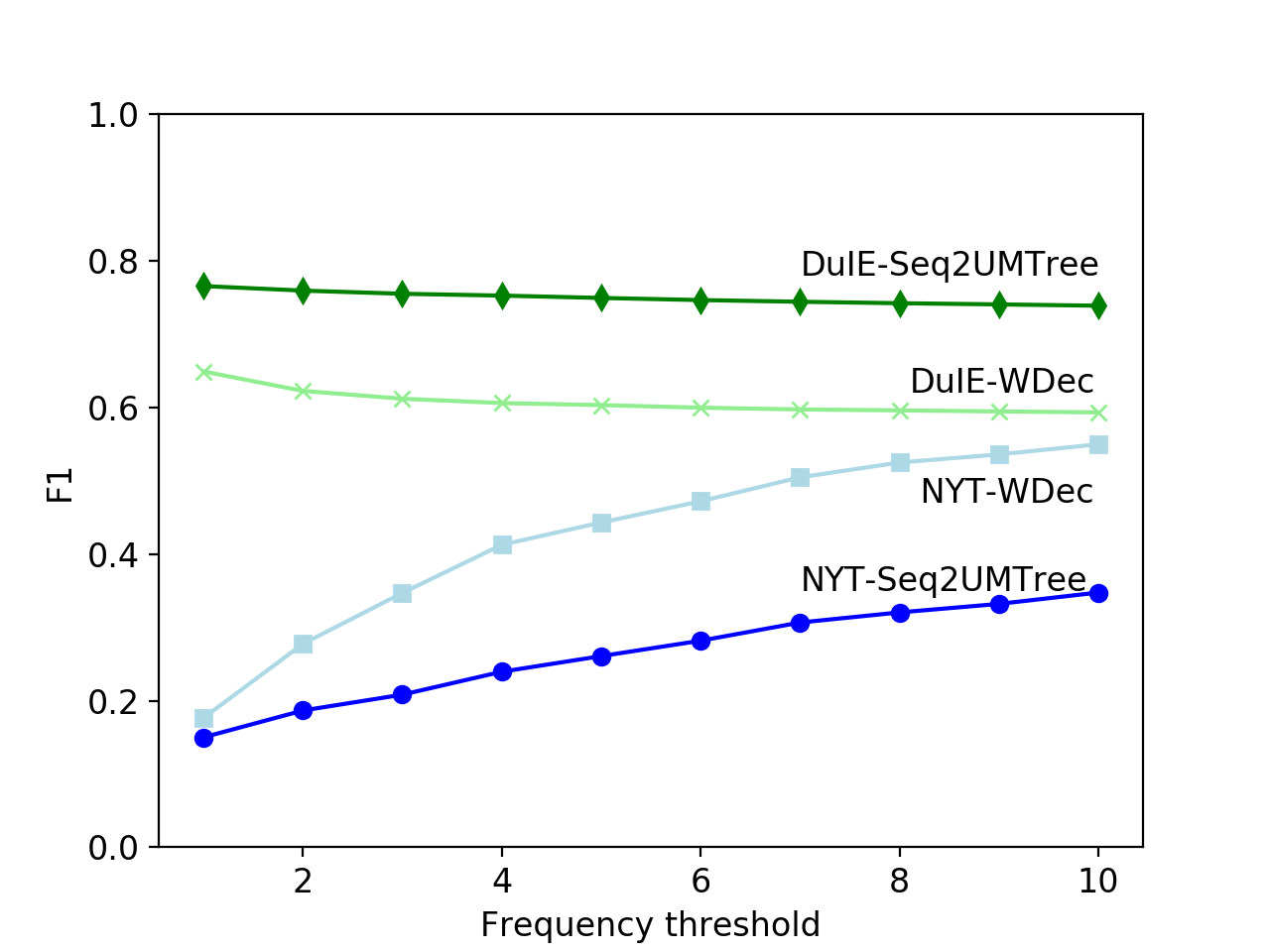}
  \caption{The F1 scores of the models with triplet frequency less than threshold. Triplet frequency represents how often the test triplets appeared in the training set.}
  \label{fig:overlap}
  \end{figure}

To investigate the effects of data bias from reoccurred frequency,
we split the test set into 10 subsets according to the reoccurred frequency (1-10) of triplets in the training set. 
The results are shown in Figure. \ref{fig:overlap}. In NYT, the F1 scores of both WDec and Seq2UMTree increases as the reoccurred frequency increases. In DuIE, the performance curve is almost flat despite of the reoccurred frequency. 
This implies that the performance is highly related to the reoccurred frequency in NYT (90\% reoccurred) but is minimally related to that in DuIE (30\% reoccurred). 

To further testify the effects of exposure bias on seen and unseen data,
we conduct an AB test on the NYT dataset. 
We take a new training set from the NYT training set, and then take two new test sets, Test-A and Test-B, from the NYT test set: Test-A's triplets is 100\% overlapped with these in the new training set but the triplets in Test-B have never appeared in the new training set. 
The new training set consists of 60\% of the original. Test-A and Test-B contain 53\% and 47\% of the sentences from the original, respectively. The results are reported in Table. \ref{tab:abtest}. 

Though Seq2UMTree underperforms WDec in 100\%-overlapped set, it outperforms WDec in unseen set. The performance drop (from seen to unseen) for Seq2UMTree is smaller than WDec's, which implies that Seq2UMTree is more robust and more reliable.
This verifies our hypothesis that the Seq2Seq models suffer more from exposure bias, which results in more overfitting, while Seq2UMtree with minimized exposure bias is more generalized to the unseen triplets.


\begin{table}[t]
\normalsize
\centering
\begin{tabular}{ll|ccc}
\hline
                      &            & Prec & Rec  & F1   \\ \hline
\multirow{2}{*}{Test-A} & Seq2UMTree & .891 & .882 & .886 \\
                      & WDec       & .956 & .862 & .906 \\ \hline
\multirow{2}{*}{Test-B}  & Seq2UMTree & .695 & .579 & .631 \\
                      & WDec       & .616 & .562 & .588 \\ \hline
\end{tabular}
\caption{AB-Test on NYT. We split NYT test set to two subsets. The triplets in Test-A set (2,625 sentences) 100\% have occurred in the filtered training subset (33,963 sentences) while the triplets in Test-B set (2,317 sentences) have never occurred in the filtered training set.}
\label{tab:abtest}
\end{table}



As the NYT dataset intrinsically has high portion of overlapped triplets in its training and test sets, and has already been overfitted by existing models, we suggest that NYT is not unbiased enough to be used as a baseline dataset, and the F1 scores of the models on NYT are not reliable.



\begin{table}[t]
\normalsize
\centering
\begin{tabular}{l|cccc}
\hline
                      & Order   & Prec          & Rec           & F1            \\ \hline
\multirow{6}{*}{NYT}  & t, r, h & .788          & .694          & .738          \\
                      & r, t, h & .791          & \textbf{.751} & \textbf{.771} \\
                      & t, h, r & .765          & .495          & .601          \\
                      & h, t, r & .756          & .548          & .635          \\
                      & r, h, t & .789          & .737          & .762          \\
                      & h, r, t & \textbf{.796} & .685          & .737          \\ \hline
\multirow{6}{*}{DuIE} & t, r, h & .766          & .663          & .711          \\
                      & r, t, h & .756          & \textbf{.730} & \textbf{.743} \\
                      & t, h, r & \textbf{.802} & .330          & .467          \\
                      & h, t, r & .794          & .120          & .208          \\
                      & r, h, t & .760          & .712          & .735          \\
                      & h, r, t & .731          & .728          & .729          \\ \hline
\end{tabular}
\caption{Different orders of Seq2UMTree.} 
\label{tab:order}
\end{table}

\subsection{Orders within Triplets}


\label{sec:order}


In Seq2UMTree, the relation, head entity and tail entity are still decoded in a predefined order (e.g., $h$-$r$-$t$ or $r$-$t$-$h$). We enumerate all six possible decoding orders in each dataset and compare the performances. The results are shown in Table. \ref{tab:order}. The performances varies by order within triplets, while the recalls for orders $t$-$h$-$r$ and $h$-$t$-$r$ drop drastically in both datasets, respectively.


We then hypothesize that the order within the triplets matters in some way. Thinking of this, we decide to look into the training phase time step by time step, and find that these 2 orders cannot even fit training set well: the recall for $h$-$t$-$r$ is only 13\% on the training set (12\% on the test set). Moreover, most of the the predictions are missing on the first time step ($h$). 
This implies that the position of $r$ provides information important to the predictions and proved our hypothesis.
By thorough error analysis, we realize that for the order $h$-$t$-$r$ ($t$-$h$-$r$ follows the same logic), the model has to predict all $t$ with regard to $h$ in the second time step, without constraints from the $r$, and this makes every possible entity to be a prediction candidate. However, the model is unable to eliminate no-relation entity pairs at the third time step, thus the model is prone to feed entity pairs to the classification layer with an low odds (low recall) but high confidence (high precision).

In contrast, for the order $h$-$r$-$t$, given the predicted $h$, the corresponding $r$ can be easily identified according to the context. Subsequently, the predicted $h$-$r$ pair gives strong hint to the last time step prediction, hence the model will not collapse from the no-relation. This also applies to any other order with $r$ in the first two time steps. 
\section{Related Work}



Previous work uses \textsc{Pipeline} to extract triplets from text \cite{pipeline1,pipeline2}. They first recognize all entities in the input sentence then classify relations for each entity pair exhaustively. \citeauthor{li-ji-2014-incremental} (\citeyear{li-ji-2014-incremental}) point out that the classification errors may propagate across subtasks. 
Instead of treating these two subtasks separately, for joint entities and relations extraction (\textsc{JERE}), 
\textsc{Table} methods calculate the similarity score of
all token pairs and relations by exhaustive enumeration
and the extracted triplets are found by the position of the output in the table \cite{miwa-bansal-2016-end,gupta-etal-2016-table}.
However, as a triplet may contain entities with different lengths, the table methods either suffer from exponential computational burden \cite{adel-schutze-2017-global} or {roll back} to pipeline methods \cite{sun-etal-2018-extracting,bekoulis2018joint,fu-etal-2019-graphrel}. 
Furthermore, 
such table enumeration dilutes the positive labels quadratically, thus
aggravating the class-imbalanced problem.
To model the task in a more concise way, \newcite{zheng-etal-2017-joint} propose a \textsc{NovelTagging} scheme, which represents relation and entity in one tag, so that the joint extraction can be solved by the well-studied sequence labeling approach. However, this tagging scheme cannot assign multiple tags to one token thus fail on overlapping triplets. The follow-on methods revise the tagging scheme to enable multi-pass sequence labeling \cite{rlre,tag2table} but they introduce 
a similar sparsity issue as does the table method.

Another promising method, \textsc{Seq2Seq}, is first proposed by \newcite{zeng-etal-2018-extracting}. Seq2Seq does not only decode the triplet list straightforwardly but also circumvents the overlapping triplets problem. 
Although this paper introduces a problem
that multi-token entities cannot be predicted, this problem has been solved by
multiple follow-up papers \cite{zeng2019copymtl,nayak2019effective}.
However, there still remains a weakness in Seq2Seq models, i.e., the exposure bias, which has been overlooked.

Exposure bias originates from the discrepancy between training and testing: Seq2Seq models use data distribution for training and model distribution for testing \cite{ranzato2015sequence}. 
Existing work mainly focuses on how to mitigate the information loss of $\arg \max$ sampling \cite{yang-etal-2018-sgm,yang-etal-2019-deep,zhang-etal-2019-bridging}. 
\newcite{NIPS2017_7125} notice that different orders affect the performance of the Seq2Seq models in Multi-Class Classification (MCC), and conduct thoroughly experiments on frequency order and topology order.
In \textsc{JERE}, \newcite{zeng-etal-2019-learning} {study additional rule-based triplet prediction orders}, including alphabetical, shuffle and fix-unsort, and then propose a reinforcement learning framework to generate triplets in adaptive orders dynamically. 
\newcite{tsai2019order} first point out the unnecessary order causes exposure bias altering the performance in MCC, and they find that Seq2Seq models are prone to overfit to the frequent label combination and show poor generalization on unseen target sequence. 

Our method solves the exposure bias problem. 
As the exposure bias problem stems from the ordered left-to-right triplet decoding, 
we block the decoding of them from each other by removing the order of the triplet generation, thus the possible prediction error cannot propagate from triplet to triplet.
Furthermore, because each triplet is generated by an independent decoding process, the decoding length has been extremely shortened, thus minimizes the effects of exposure bias.
Our method differs from previous solution on exposure bias that we remove the order by structure decoding rather than random sampling \cite{tsai2019order}.

CASREL \cite{jianlin} is a recently proposed two-step tagging method, which first finds all the head entities in the sentence then labels a relation-tail table for each head entity, which can also be seen as a UMTree decoder with a decoding length two. However, they overlook the data bias problem in NYT, which causing model unreliability and possible model bias.

Note that our task is different from \textsc{ONEIE} \cite{lin2020oneie}, which models event extraction, entity span detection, entity type recognition and relation extraction in a Seq2Graph way. 
In contrast to \textsc{ONEIE}, \textsc{JERE} aims to extract only relation-entity triplets, which can be modeled by our UMTree structure naturally. 
The simplicity of the tree enables the model to conduct global extraction.


\section{Conclusions}

In this paper, we thoroughly analyze the effects of exposure bias of Seq2Seq models on joint entity and relation extraction. Exposure bias causes overfitting that hurts the reliability of the performance scores. To solve the problem of exposure bias, we point out the order of the target triplets is redundant and formulate the target triplet sequence to Unordered-Multi-Tree.
The Unordered-Multi-Tree structure minimizes the effect of exposure bias by limiting the decoding length to three within a triplet, and removing the order among triplets.
We conduct in-depth experiments and reveal the relationship between exposure bias and data bias. 
The results show great generalization of our model.


\bibliography{ref}

\begin{thebibliography}{49}
\expandafter\ifx\csname natexlab\endcsname\relax\def\natexlab#1{#1}\fi

\bibitem[{Adel and Sch{\"u}tze(2017)}]{adel-schutze-2017-global}
Heike Adel and Hinrich Sch{\"u}tze. 2017.
\newblock \href {https://doi.org/10.18653/v1/D17-1181} {Global normalization of
  convolutional neural networks for joint entity and relation classification}.
\newblock In \emph{Proceedings of the 2017 Conference on Empirical Methods in
  Natural Language Processing}, pages 1723--1729, Copenhagen, Denmark.
  Association for Computational Linguistics.

\bibitem[{Bekoulis et~al.(2018)Bekoulis, Deleu, Demeester, and
  Develder}]{bekoulis2018joint}
Giannis Bekoulis, Johannes Deleu, Thomas Demeester, and Chris Develder. 2018.
\newblock Joint entity recognition and relation extraction as a multi-head
  selection problem.
\newblock \emph{Expert Systems with Applications}, 114:34--45.

\bibitem[{Benmalek et~al.(2019)Benmalek, Khabsa, Desu, Cardie, and
  Banko}]{benmalek-etal-2019-keeping}
Ryan Benmalek, Madian Khabsa, Suma Desu, Claire Cardie, and Michele Banko.
  2019.
\newblock \href {https://doi.org/10.18653/v1/P19-1407} {Keeping notes:
  Conditional natural language generation with a scratchpad encoder}.
\newblock In \emph{Proceedings of the 57th Annual Meeting of the Association
  for Computational Linguistics}, pages 4157--4167, Florence, Italy.
  Association for Computational Linguistics.

\bibitem[{Chan and Roth(2011)}]{pipeline2}
Yee~Seng Chan and Dan Roth. 2011.
\newblock \href {https://www.aclweb.org/anthology/P11-1056} {Exploiting
  syntactico-semantic structures for relation extraction}.
\newblock In \emph{Proceedings of ACL}, pages 551--560, Portland, Oregon, USA.
  Association for Computational Linguistics.

\bibitem[{Chen et~al.(2019)Chen, Yuan, Wang, and Bai}]{chen-etal-2019-mrmep}
Jiayu Chen, Caixia Yuan, Xiaojie Wang, and Ziwei Bai. 2019.
\newblock \href {https://doi.org/10.18653/v1/K19-1055} {{M}r{M}ep: Joint
  extraction of multiple relations and multiple entity pairs based on triplet
  attention}.
\newblock In \emph{Proceedings of the 23rd Conference on Computational Natural
  Language Learning (CoNLL)}, pages 593--602, Hong Kong, China. Association for
  Computational Linguistics.

\bibitem[{Cho et~al.(2014)Cho, van Merri{\"e}nboer, Gulcehre, Bahdanau,
  Bougares, Schwenk, and Bengio}]{cho-etal-2014-learning}
Kyunghyun Cho, Bart van Merri{\"e}nboer, Caglar Gulcehre, Dzmitry Bahdanau,
  Fethi Bougares, Holger Schwenk, and Yoshua Bengio. 2014.
\newblock \href {https://doi.org/10.3115/v1/D14-1179} {Learning phrase
  representations using {RNN} encoder{--}decoder for statistical machine
  translation}.
\newblock In \emph{Proceedings of the 2014 Conference on Empirical Methods in
  Natural Language Processing ({EMNLP})}, pages 1724--1734, Doha, Qatar.
  Association for Computational Linguistics.

\bibitem[{Dai et~al.(2019)Dai, Xiao, Lyu, Dou, She, and Wang}]{tag2table}
Dai Dai, Xinyan Xiao, Yajuan Lyu, Shan Dou, Qiaoqiao She, and Haifeng Wang.
  2019.
\newblock \href {https://doi.org/10.1609/aaai.v33i01.33016300} {Joint
  extraction of entities and overlapping relations using position-attentive
  sequence labeling}.
\newblock \emph{Proceedings of AAAI}, 33(01):6300--6308.

\bibitem[{Dong and Lapata(2016)}]{dong-lapata-2016-language}
Li~Dong and Mirella Lapata. 2016.
\newblock \href {https://doi.org/10.18653/v1/P16-1004} {Language to logical
  form with neural attention}.
\newblock In \emph{Proceedings of the 54th Annual Meeting of the Association
  for Computational Linguistics (Volume 1: Long Papers)}, pages 33--43, Berlin,
  Germany. Association for Computational Linguistics.

\bibitem[{Durrett and Klein(2014)}]{Durrett2014}
Greg Durrett and Dan Klein. 2014.
\newblock A joint model for entity analysis: Coreference, typing, and linking.
\newblock In \emph{Transactions of the Association for Computational
  Linguistics (TACL)}.

\bibitem[{Fu et~al.(2019)Fu, Li, and Ma}]{fu-etal-2019-graphrel}
Tsu-Jui Fu, Peng-Hsuan Li, and Wei-Yun Ma. 2019.
\newblock \href {https://doi.org/10.18653/v1/P19-1136} {{G}raph{R}el: Modeling
  text as relational graphs for joint entity and relation extraction}.
\newblock In \emph{Proceedings of the 57th Annual Meeting of the Association
  for Computational Linguistics}, pages 1409--1418, Florence, Italy.
  Association for Computational Linguistics.

\bibitem[{Gupta et~al.(2016)Gupta, Sch{\"u}tze, and
  Andrassy}]{gupta-etal-2016-table}
Pankaj Gupta, Hinrich Sch{\"u}tze, and Bernt Andrassy. 2016.
\newblock \href {https://www.aclweb.org/anthology/C16-1239} {Table filling
  multi-task recurrent neural network for joint entity and relation
  extraction}.
\newblock In \emph{Proceedings of {COLING} 2016, the 26th International
  Conference on Computational Linguistics: Technical Papers}, pages 2537--2547,
  Osaka, Japan. The COLING 2016 Organizing Committee.

\bibitem[{Hochreiter and Schmidhuber(1997)}]{lstm}
Sepp Hochreiter and Jürgen Schmidhuber. 1997.
\newblock \href {https://doi.org/10.1162/neco.1997.9.8.1735} {Long short-term
  memory}.
\newblock \emph{Neural Computation}, 9(8):1735--1780.

\bibitem[{Ji and Grishman(2005)}]{Ji2005a}
Heng Ji and Ralph Grishman. 2005.
\newblock Improving name tagging by reference resolution and relation
  detection.
\newblock In \emph{In Proceedings of ACL 05, Ann Arbor, USA}.

\bibitem[{Ji et~al.(2005)Ji, Westbrook, and Grishman}]{Ji2005b}
Heng Ji, David Westbrook, and Ralph Grishman. 2005.
\newblock Using semantic relations to refine coreference decisions.
\newblock In \emph{In Proceedings of HLT/EMNLP 05, Vancouver, B.C., Canada}.

\bibitem[{Katiyar and Cardie(2017)}]{limb}
Arzoo Katiyar and Claire Cardie. 2017.
\newblock \href {https://doi.org/10.18653/v1/P17-1085} {Going out on a limb:
  Joint extraction of entity mentions and relations without dependency trees}.
\newblock In \emph{Proceedings of ACL}, pages 917--928, Vancouver, Canada.
  Association for Computational Linguistics.

\bibitem[{Kingma and Ba(2014)}]{kingma2014method}
Diederik~P. Kingma and Jimmy Ba. 2014.
\newblock \href {http://arxiv.org/abs/1412.6980} {Adam: A method for stochastic
  optimization}.
\newblock Cite arxiv:1412.6980Comment: Published as a conference paper at the
  3rd International Conference for Learning Representations, San Diego, 2015.

\bibitem[{Kirschnick et~al.(2016)Kirschnick, Hemsen, and
  Markl}]{kirschnick-etal-2016-jedi}
Johannes Kirschnick, Holmer Hemsen, and Volker Markl. 2016.
\newblock {JEDI}: Joint entity and relation detection using type inference.
\newblock In \emph{Proceedings of the 54th Annual Meeting of the Association
  for Computational Linguistics System Demonstrations (ACL2016)}.

\bibitem[{Li and Ji(2014)}]{li-ji-2014-incremental}
Qi~Li and Heng Ji. 2014.
\newblock \href {https://doi.org/10.3115/v1/P14-1038} {Incremental joint
  extraction of entity mentions and relations}.
\newblock In \emph{Proceedings of the 52nd Annual Meeting of the Association
  for Computational Linguistics (Volume 1: Long Papers)}, pages 402--412,
  Baltimore, Maryland. Association for Computational Linguistics.

\bibitem[{Li et~al.(2014)Li, Ji, Yu, and Li}]{li2014constructing}
Qi~Li, Heng Ji, HONG Yu, and Sujian Li. 2014.
\newblock Constructing information networks using one single model.
\newblock In \emph{Proceedings of the 2014 Conference on Empirical Methods in
  Natural Language Processing (EMNLP2014)}.

\bibitem[{Li et~al.(2019)Li, He, Shi, Jiang, Liang, Jiang, Zhang, Lyu, and
  Zhu}]{li2019duie}
Shuangjie Li, Wei He, Yabing Shi, Wenbin Jiang, Haijin Liang, Ye~Jiang, Yang
  Zhang, Yajuan Lyu, and Yong Zhu. 2019.
\newblock Duie: A large-scale chinese dataset for information extraction.
\newblock In \emph{CCF International Conference on Natural Language Processing
  and Chinese Computing}, pages 791--800. Springer.

\bibitem[{Lin et~al.(2020)Lin, Ji, Huang, and Wu}]{lin2020oneie}
Ying Lin, Heng Ji, Fei Huang, and Lingfei Wu. 2020.
\newblock A joint end-to-end neural model for information extraction with
  global features.
\newblock In \emph{Proceedings of The 58th Annual Meeting of the Association
  for Computational Linguistics}.

\bibitem[{Liu et~al.(2019)Liu, Guan, Li, and Kawahara}]{liu-etal-2019-tree}
Qianying Liu, Wenyv Guan, Sujian Li, and Daisuke Kawahara. 2019.
\newblock \href {https://doi.org/10.18653/v1/D19-1241} {Tree-structured
  decoding for solving math word problems}.
\newblock In \emph{Proceedings of the 2019 Conference on Empirical Methods in
  Natural Language Processing and the 9th International Joint Conference on
  Natural Language Processing (EMNLP-IJCNLP)}, pages 2370--2379, Hong Kong,
  China. Association for Computational Linguistics.

\bibitem[{Lu and Roth(2015)}]{lu-roth-2015-joint}
Wei Lu and Dan Roth. 2015.
\newblock Joint mention extraction and classification with mention hypergraphs.
\newblock In \emph{Proceedings of the 2015 Conference on Empirical Methods in
  Natural Language Processing (EMNLP2015)}.

\bibitem[{Luong et~al.(2015)Luong, Pham, and
  Manning}]{luong-etal-2015-effective}
Thang Luong, Hieu Pham, and Christopher~D. Manning. 2015.
\newblock \href {https://doi.org/10.18653/v1/D15-1166} {Effective approaches to
  attention-based neural machine translation}.
\newblock In \emph{Proceedings of the 2015 Conference on Empirical Methods in
  Natural Language Processing}, pages 1412--1421, Lisbon, Portugal. Association
  for Computational Linguistics.

\bibitem[{Miwa and Bansal(2016)}]{miwa-bansal-2016-end}
Makoto Miwa and Mohit Bansal. 2016.
\newblock \href {https://doi.org/10.18653/v1/P16-1105} {End-to-end relation
  extraction using {LSTM}s on sequences and tree structures}.
\newblock In \emph{Proceedings of the 54th Annual Meeting of the Association
  for Computational Linguistics (Volume 1: Long Papers)}, pages 1105--1116,
  Berlin, Germany. Association for Computational Linguistics.

\bibitem[{Miwa and Sasaki(2014)}]{miwa-sasaki-2014-modeling}
Makoto Miwa and Yutaka Sasaki. 2014.
\newblock \href {https://doi.org/10.3115/v1/D14-1200} {Modeling joint entity
  and relation extraction with table representation}.
\newblock In \emph{Proceedings of the 2014 Conference on Empirical Methods in
  Natural Language Processing ({EMNLP})}, pages 1858--1869, Doha, Qatar.
  Association for Computational Linguistics.

\bibitem[{Nadeau and Sekine(2007)}]{pipeline1}
David Nadeau and Satoshi Sekine. 2007.
\newblock A survey of named entity recognition and classification.
\newblock \emph{Lingvisticae Investigationes}, 30(1):3--26.

\bibitem[{Nam et~al.(2017)Nam, Loza~Menc\'{\i}a, Kim, and
  F\"{u}rnkranz}]{NIPS2017_7125}
Jinseok Nam, Eneldo Loza~Menc\'{\i}a, Hyunwoo~J Kim, and Johannes
  F\"{u}rnkranz. 2017.
\newblock \href
  {http://papers.nips.cc/paper/7125-maximizing-subset-accuracy-with-recurrent-neural-networks-in-multi-label-classification.pdf}
  {Maximizing subset accuracy with recurrent neural networks in multi-label
  classification}.
\newblock In I.~Guyon, U.~V. Luxburg, S.~Bengio, H.~Wallach, R.~Fergus,
  S.~Vishwanathan, and R.~Garnett, editors, \emph{Advances in Neural
  Information Processing Systems 30}, pages 5413--5423. Curran Associates, Inc.

\bibitem[{Nayak and Ng(2019)}]{nayak2019effective}
Tapas Nayak and Hwee~Tou Ng. 2019.
\newblock Effective modeling of encoder-decoder architecture for joint entity
  and relation extraction.
\newblock \emph{arXiv preprint arXiv:1911.09886}.

\bibitem[{Ranzato et~al.(2015)Ranzato, Chopra, Auli, and
  Zaremba}]{ranzato2015sequence}
Marc'Aurelio Ranzato, Sumit Chopra, Michael Auli, and Wojciech Zaremba. 2015.
\newblock Sequence level training with recurrent neural networks.
\newblock \emph{arXiv preprint arXiv:1511.06732}.

\bibitem[{Riedel et~al.(2010)Riedel, Yao, and McCallum}]{riedel2010modeling}
Sebastian Riedel, Limin Yao, and Andrew McCallum. 2010.
\newblock Modeling relations and their mentions without labeled text.
\newblock In \emph{Joint European Conference on Machine Learning and Knowledge
  Discovery in Databases}, pages 148--163. Springer.

\bibitem[{Roth and Yih(2004)}]{Roth2004}
Dan Roth and Wen-tau Yih. 2004.
\newblock A linear programming formulation for global inference in natural
  language tasks.
\newblock In \emph{Proceedings of the Eighth Conference on Computational
  Natural Language Learning (CoNLL2004)}.

\bibitem[{{Schuster} and {Paliwal}(1997)}]{birnn}
M.~{Schuster} and K.~K. {Paliwal}. 1997.
\newblock Bidirectional recurrent neural networks.
\newblock \emph{IEEE Transactions on Signal Processing}, 45(11):2673--2681.

\bibitem[{Sil and Yates(2013)}]{Sil2013}
Avirup Sil and Alexander Yates. 2013.
\newblock Re-ranking for joint named-entity recognition and linking.
\newblock In \emph{Proceedings of the 22nd ACM international conference on
  Conference on Information \& Knowledge Management (CIKM2013)}.

\bibitem[{Sun et~al.(2018)Sun, Wu, Lan, Sun, Wang, Lee, and
  Wu}]{sun-etal-2018-extracting}
Changzhi Sun, Yuanbin Wu, Man Lan, Shiliang Sun, Wenting Wang, Kuang-Chih Lee,
  and Kewen Wu. 2018.
\newblock \href {https://doi.org/10.18653/v1/D18-1249} {Extracting entities and
  relations with joint minimum risk training}.
\newblock In \emph{Proceedings of the 2018 Conference on Empirical Methods in
  Natural Language Processing}, pages 2256--2265, Brussels, Belgium.
  Association for Computational Linguistics.

\bibitem[{Takanobu et~al.(2018)Takanobu, Zhang, Liu, and Huang}]{rlre}
Ryuichi Takanobu, Tianyang Zhang, Jiexi Liu, and Minlie Huang. 2018.
\newblock A hierarchical framework for relation extraction with reinforcement
  learning.
\newblock \emph{arXiv preprint arXiv:1811.03925}.

\bibitem[{Tsai and Lee(2019)}]{tsai2019order}
Che-Ping Tsai and Hung-Yi Lee. 2019.
\newblock Order-free learning alleviating exposure bias in multi-label
  classification.
\newblock \emph{arXiv preprint arXiv:1909.03434}.

\bibitem[{Wei et~al.(2019)Wei, Su, Wang, Tian, and Chang}]{wei2019novel}
Zhepei Wei, Jianlin Su, Yue Wang, Yuan Tian, and Yi~Chang. 2019.
\newblock A novel hierarchical binary tagging framework for joint extraction of
  entities and relations.
\newblock \emph{arXiv preprint arXiv:1909.03227}.

\bibitem[{Wei et~al.(2020)Wei, Su, Wang, Tian, and Chang}]{jianlin}
Zhepei Wei, Jianlin Su, Yue Wang, Yuan Tian, and Yi~Chang. 2020.
\newblock \href {https://doi.org/10.18653/v1/2020.acl-main.136} {A novel
  cascade binary tagging framework for relational triple extraction}.
\newblock In \emph{Proceedings of the 58th Annual Meeting of the Association
  for Computational Linguistics}, pages 1476--1488, Online. Association for
  Computational Linguistics.

\bibitem[{{Williams} and {Zipser}(1989)}]{6795228}
R.~J. {Williams} and D.~{Zipser}. 1989.
\newblock A learning algorithm for continually running fully recurrent neural
  networks.
\newblock \emph{Neural Computation}, 1(2):270--280.

\bibitem[{Yang and Mitchell(2016)}]{yang-mitchell-2016-joint}
Bishan Yang and Tom~M. Mitchell. 2016.
\newblock Joint extraction of events and entities within a document context.
\newblock In \emph{Proceedings of the 2016 Conference of the North American
  Chapter of the Association for Computational Linguistics: Human Language
  Technologies (NAACL HLT2016)}.

\bibitem[{Yang et~al.(2019)Yang, Luo, Ma, Lin, and Sun}]{yang-etal-2019-deep}
Pengcheng Yang, Fuli Luo, Shuming Ma, Junyang Lin, and Xu~Sun. 2019.
\newblock \href {https://doi.org/10.18653/v1/P19-1518} {A deep reinforced
  sequence-to-set model for multi-label classification}.
\newblock In \emph{Proceedings of the 57th Annual Meeting of the Association
  for Computational Linguistics}, pages 5252--5258, Florence, Italy.
  Association for Computational Linguistics.

\bibitem[{Yang et~al.(2018)Yang, Sun, Li, Ma, Wu, and
  Wang}]{yang-etal-2018-sgm}
Pengcheng Yang, Xu~Sun, Wei Li, Shuming Ma, Wei Wu, and Houfeng Wang. 2018.
\newblock \href {https://www.aclweb.org/anthology/C18-1330} {{SGM}: Sequence
  generation model for multi-label classification}.
\newblock In \emph{Proceedings of the 27th International Conference on
  Computational Linguistics}, pages 3915--3926, Santa Fe, New Mexico, USA.
  Association for Computational Linguistics.

\bibitem[{Yu and Lam(2010)}]{yu-lam-2010-jointly}
Xiaofeng Yu and Wai Lam. 2010.
\newblock \href {https://www.aclweb.org/anthology/C10-2160} {Jointly
  identifying entities and extracting relations in encyclopedia text via a
  graphical model approach}.
\newblock In \emph{Coling 2010: Posters}, pages 1399--1407, Beijing, China.
  Coling 2010 Organizing Committee.

\bibitem[{Zeng et~al.(2019{\natexlab{a}})Zeng, Zhang, and
  Liu}]{zeng2019copymtl}
Daojian Zeng, Haoran Zhang, and Qianying Liu. 2019{\natexlab{a}}.
\newblock Copymtl: Copy mechanism for joint extraction of entities and
  relations with multi-task learning.
\newblock \emph{arXiv preprint arXiv:1911.10438}.

\bibitem[{Zeng et~al.(2019{\natexlab{b}})Zeng, He, Zeng, Liu, Liu, and
  Zhao}]{zeng-etal-2019-learning}
Xiangrong Zeng, Shizhu He, Daojian Zeng, Kang Liu, Shengping Liu, and Jun Zhao.
  2019{\natexlab{b}}.
\newblock \href {https://doi.org/10.18653/v1/D19-1035} {Learning the extraction
  order of multiple relational facts in a sentence with reinforcement
  learning}.
\newblock In \emph{Proceedings of the 2019 Conference on Empirical Methods in
  Natural Language Processing and the 9th International Joint Conference on
  Natural Language Processing (EMNLP-IJCNLP)}, pages 367--377, Hong Kong,
  China. Association for Computational Linguistics.

\bibitem[{Zeng et~al.(2018)Zeng, Zeng, He, Liu, and
  Zhao}]{zeng-etal-2018-extracting}
Xiangrong Zeng, Daojian Zeng, Shizhu He, Kang Liu, and Jun Zhao. 2018.
\newblock \href {https://doi.org/10.18653/v1/P18-1047} {Extracting relational
  facts by an end-to-end neural model with copy mechanism}.
\newblock In \emph{Proceedings of the 56th Annual Meeting of the Association
  for Computational Linguistics (Volume 1: Long Papers)}, pages 506--514,
  Melbourne, Australia. Association for Computational Linguistics.

\bibitem[{Zhang et~al.(2019)Zhang, Feng, Meng, You, and
  Liu}]{zhang-etal-2019-bridging}
Wen Zhang, Yang Feng, Fandong Meng, Di~You, and Qun Liu. 2019.
\newblock \href {https://doi.org/10.18653/v1/P19-1426} {Bridging the gap
  between training and inference for neural machine translation}.
\newblock In \emph{Proceedings of the 57th Annual Meeting of the Association
  for Computational Linguistics}, pages 4334--4343, Florence, Italy.
  Association for Computational Linguistics.

\bibitem[{Zheng et~al.(2017)Zheng, Wang, Bao, Hao, Zhou, and
  Xu}]{zheng-etal-2017-joint}
Suncong Zheng, Feng Wang, Hongyun Bao, Yuexing Hao, Peng Zhou, and Bo~Xu. 2017.
\newblock \href {https://doi.org/10.18653/v1/P17-1113} {Joint extraction of
  entities and relations based on a novel tagging scheme}.
\newblock In \emph{Proceedings of the 55th Annual Meeting of the Association
  for Computational Linguistics (Volume 1: Long Papers)}, pages 1227--1236,
  Vancouver, Canada. Association for Computational Linguistics.

\end{thebibliography}
\bibliographystyle{acl_natbib}

\end{document}